\pdfoutput=1

\documentclass[11pt]{article}

\usepackage[preprint]{acl}

\usepackage{times}
\usepackage{latexsym}
\usepackage{amsmath}
\usepackage{amssymb}
\usepackage{multirow}

\usepackage[T1]{fontenc}

\usepackage[utf8]{inputenc}

\usepackage{microtype}
\usepackage{tcolorbox}
\usepackage{float}
\usepackage{subcaption}

\usepackage{inconsolata}

\usepackage{graphicx}
\usepackage{booktabs}

\title{Fin-ExBERT: User Intent based Text Extraction in Financial \\Context using Graph-Augmented BERT and trainable Plugin}

\author{Soumick Sarker \\ College of Computing and Data Science\\ Nanyang Technological University\\ Singapore\\ \texttt{soumicksarker9@gmail.com}
        \And  
        Abhijit Kumar Rai \\ Fidelity Investments\\ Bengaluru, India \\ \texttt{abhijit7000@gmail.com}}

\begin{document}
{\makeatletter\acl@finalcopytrue
  \maketitle
}
\begin{abstract}
Financial dialogue transcripts pose a unique challenge for sentence-level information extraction due to their informal structure, domain-specific vocabulary, and variable intent density. We introduce Fin-ExBERT, a lightweight and modular framework for extracting user intent–relevant sentences from annotated financial service calls. Our approach builds on a domain-adapted BERT (Bidirectional Encoder Representations from Transformers) backbone enhanced with LoRA (Low-Rank Adaptation) adapters, enabling efficient fine-tuning using limited labeled data. We propose a two-stage training strategy with progressive unfreezing: initially training a classifier head while freezing the backbone, followed by gradual fine-tuning of the entire model with differential learning rates. To ensure robust extraction under uncertainty, we adopt a dynamic thresholding strategy based on probability curvature (elbow detection), avoiding fixed cutoff heuristics.  Empirical results show strong precision and F1 performance on real-world transcripts, with interpretable output suitable for downstream auditing and question-answering workflows. The full framework supports batched evaluation, visualization, and calibrated export, offering a deployable solution for financial dialogue mining.
\end{abstract}

\section{Introduction}

Extractive text operations have become indispensable in modern industries where large volumes of unstructured textual data, such as documents, emails, and call transcripts, need to be processed efficiently to extract meaningful insights. In customer service, particularly within financial institutions, accurate text extraction plays a critical role in identifying customer queries, resolving issues, and providing personalized services. For example, identifying the context and extracting relevant spans of information from call transcripts can significantly enhance response accuracy and reduce manual effort. Such capabilities not only improve customer satisfaction but also streamline operations and reduce costs for organizations across various domains. Despite the importance of extractive text operations, achieving high accuracy in financial contexts is particularly challenging due to the domain-specific nature of financial terminology. Traditional keyword-based extraction methods often struggle to correctly interpret terms such as \textit{401k} (retirement planning) and \textit{529} (college planning) since these concepts do not always have direct linguistic correlations. Furthermore, financial texts frequently contain implicit meanings and specialized jargon that general-purpose models fail to recognize. This challenge is further exacerbated by the necessity of preserving contextual relationships across multiple utterances in multi-turn dialogues, making conventional models inadequate.

\section{Related Work}

Recent advancements in natural language processing have introduced various models tailored for financial applications, yet significant gaps remain in extractive capabilities. FinBERT~\cite{yang2020finbert}, a domain-specific adaptation of BERT~\cite{devlin2018bert}, has been successfully applied to sentiment analysis and named entity recognition tasks but lacks the necessary extractive mechanisms to handle complex multi-turn dialogues. Similarly, FinGPT~\cite{yang2023fingpt, zhang2023instructfingpt} leverages instruction-tuned LLMs for financial sentiment analysis but does not focus on extractive question-answering, making it unsuitable for tasks requiring precise span retrieval and contextual grounding.

\begin{figure*}[ht]
    \centering
    \includegraphics[width=\textwidth]{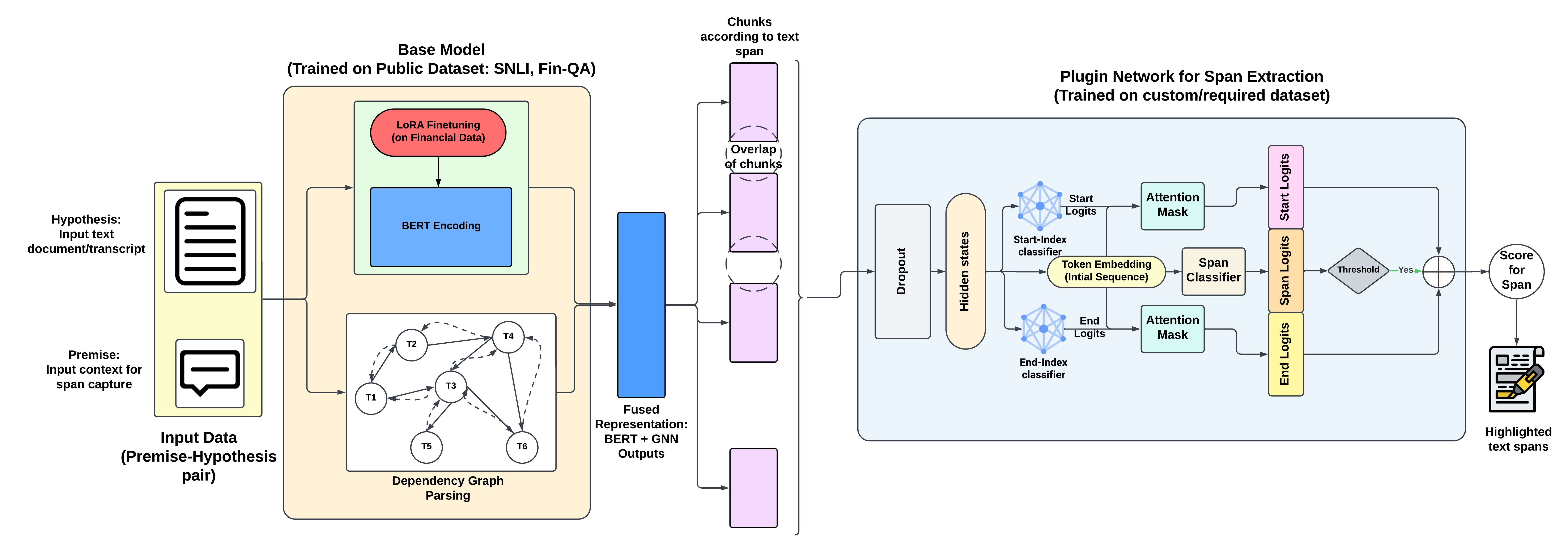}
    \caption{Flowchart of the proposed methodology showing the Base Model and the Plugin Network for text span extraction.}
    \label{fig:methodology_flowchart}
\end{figure*}

General-purpose LLMs such as GPT-4~\cite{achiam2023gpt}, LLaMA~\cite{touvron2023llama}, and PaLM~\cite{chowdhery2023palm} excel in open-domain comprehension and generation. However, their limitations in financial extraction tasks are notable: (i) high parameter counts hinder efficient fine-tuning, (ii) hallucinations~\cite{ji2023survey, ji2023towards} compromise reliability, and (iii) instruction tuning remains insufficient for domain-specific reasoning.

Several works have aimed to bridge these gaps. DeBERTa-based solvers~\cite{luo2024legal} improve contextual encoding but lack structural awareness. MT2Net~\cite{zhao2022multihiertt} and ConvFinQA~\cite{chen2022convfinqa} bring hierarchical and conversational improvements, yet fall short in technical Question-Answering(QA). DocFinQA~\cite{reddy2024docfinqa} emphasizes document reasoning over tabular inputs, while PlanGEN~\cite{parmar2025plangen} enhances logical consistency without addressing dialogue complexity. FiD~\cite{izacard2020leveraging} and KECP~\cite{wang2022kecp} offer multi-source fusion and knowledge control but underperform in noisy, domain-specific extractions.

To address these shortcomings, we propose Fin-ExBERT, a GNN-augmented BERT model tailored for extractive financial QA. It integrates syntactic reasoning via Graph Neural Networks (GNNs), LoRA adapters~\cite{hu2021lora} for lightweight domain tuning, and a tunable plugin head for scoring relevant spans. This design enables precise, scalable extraction from multi-turn transcripts and complex financial documents, advancing the robustness and interpretability of domain-specific QA systems.

\section{Methodology}
The proposed methodology integrates a Graph Neural Network (GNN) with BERT, incorporating a LoRA (Low-Rank Adaptation) adapter to improve performance on financial problem-solving tasks, along with a network for task specific text extraction. The key components of the architecture include a BERT encoder for contextual embeddings, GNNs for graph-structured reasoning, and LoRA-based domain adaptation to efficiently handle financial data. Fin-ExBERT can be divided into three stages: 
(1) Base model, which is a modified Masked Language Model (MLM), was trained using BERT and a graph augmentation for contextual and relational reasoning,
(2) Domain-specific fine-tuning using LoRA, and
(3) A trainable plugin network to extract target specific context from text. The entire flowchart is illustrated by Figure~\ref{fig:methodology_flowchart}.

\subsection{Base Model: GNN-Augmented BERT for Natural Language Inference}

\begin{figure*}[ht]
    \centering
    \includegraphics[width=0.70\textwidth]{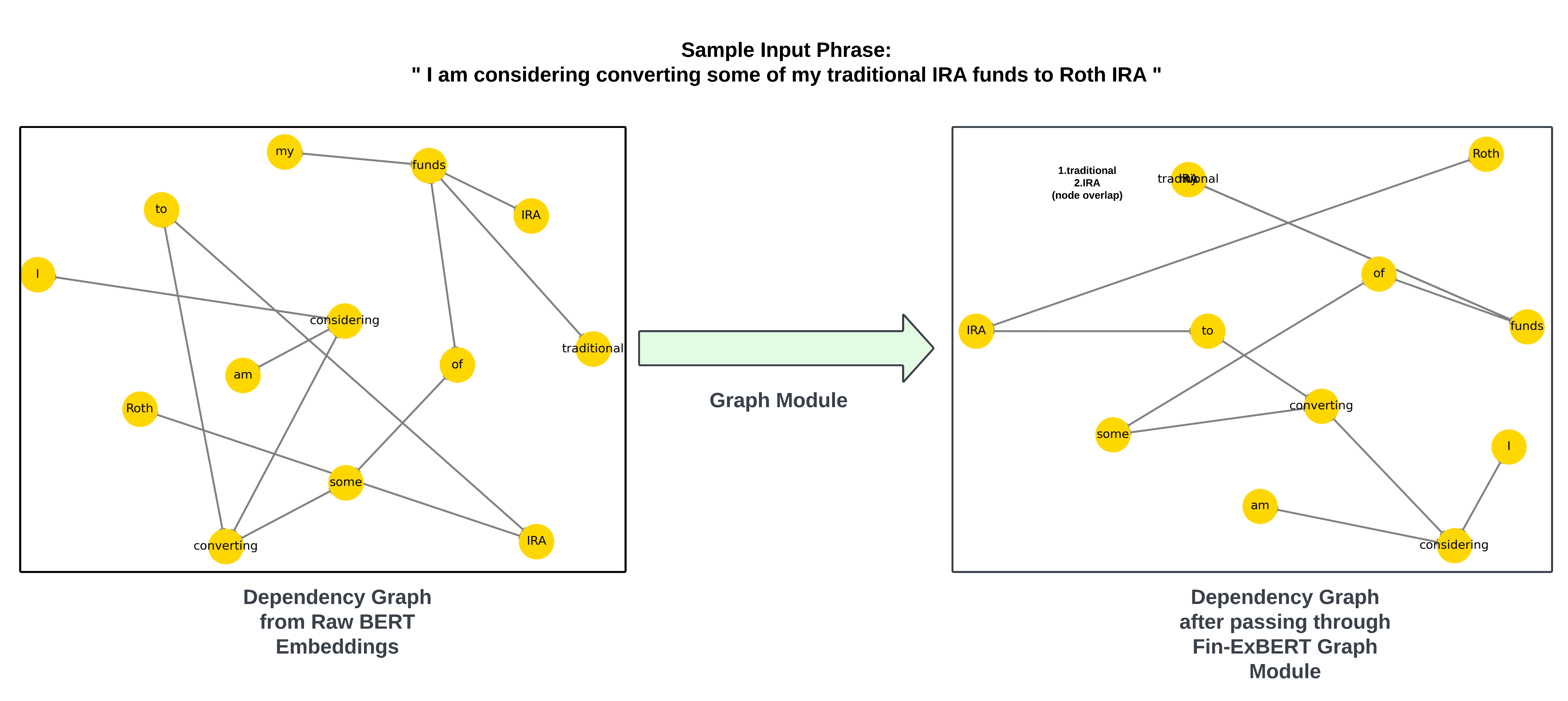}
    \caption{An illustration of how the Graph module in Fin-ExBERT modifies the dependency trees for phrases.}
    \label{fig:illustration}
\end{figure*}

\begin{figure}[ht]
    \centering
    \includegraphics[width=0.45\textwidth]{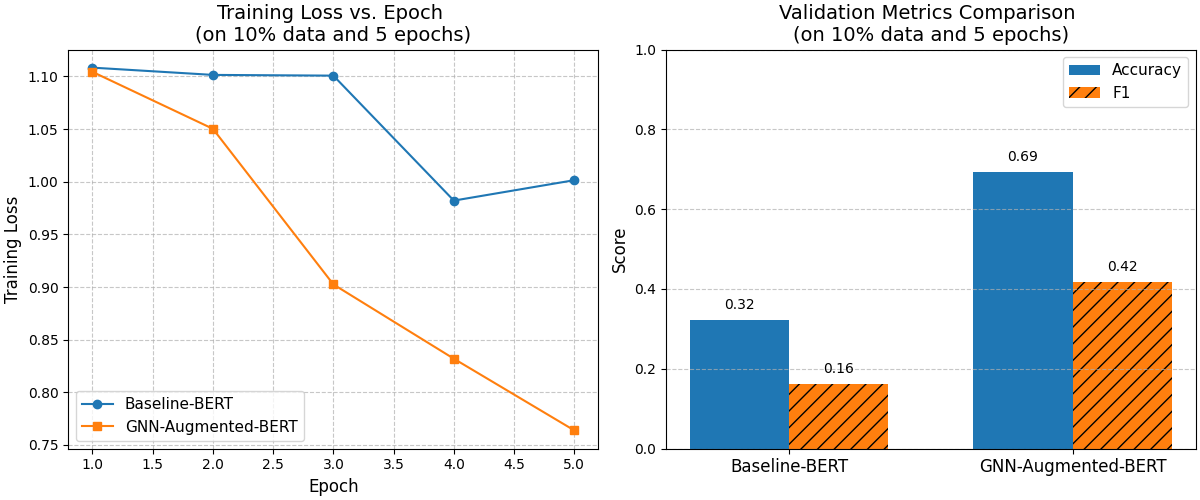}
    \caption{An ablation study illustrating the importance of the graph module.}
    \label{fig:ablation}
\end{figure}

\begin{figure}[ht]
    \centering
    \includegraphics[width=0.30\textwidth]{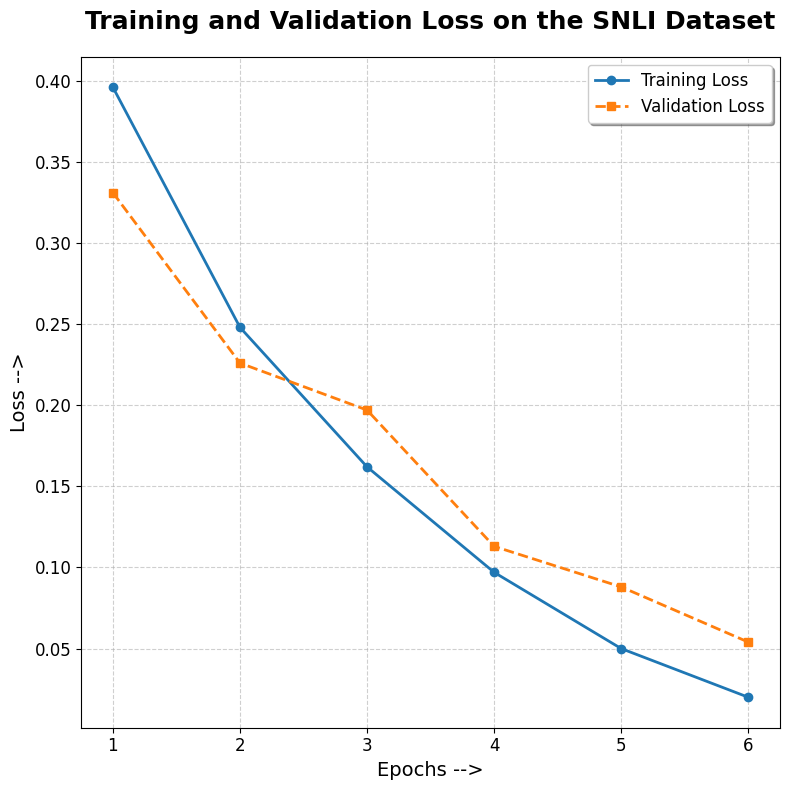}
    \caption{Loss plot while training the Base Model on the SNLI dataset.}
    \label{fig:base_model_loss_plot}
\end{figure}

The base model accepts premise–hypothesis pairs, following the standard NLI format. We use a \texttt{bert-base-uncased} encoder to obtain contextual embeddings, extracting the $[\text{CLS}]$ token as the semantic representation. To capture syntactic dependencies missed by BERT, we augment it with Graph Neural Networks (GNNs) that operate on dependency graphs generated using spaCy~\cite{spacy2}. Figure~\ref{fig:illustration} shows how the graph module modifies the dependency tree, while the importance of using a graph component here is shown by an ablation study in Figure~\ref{fig:ablation}. These graphs consist of token nodes and syntactic edges, processed through message-passing GNN layers to enhance relational reasoning. For each input pair, we extract GNN-based representations of both premise and hypothesis, and concatenate them with the $[\text{CLS}]$ (for the premise-hypothesis pair from the BERT component) embedding to form the fused representation:
\begin{equation}
\label{eq:fused_rep}
\mathbf{FR} = [\mathbf{CLS}, \mathbf{GNN}{\text{premise}}, \mathbf{GNN}{\text{hypothesis}}]
\end{equation}
This fused vector is passed to a classifier trained on the SNLI dataset~\cite{bowman2015large} for three-way NLI prediction: entailment, contradiction, and neutral. The loss plot during the training is shown in Figure~\ref{fig:base_model_loss_plot}.

\subsection{Connecting Sentence-Level Extraction to NLI-Based Pretraining}

The task of identifying relevant sentence spans in customer transcripts can be viewed as an entailment problem: given a customer utterance (hypothesis), does it entail a predefined financial intent or resolution category (premise)? Our use of NLI pretraining on the SNLI dataset allows the model to better capture such premise-hypothesis relationships, even when phrased indirectly or across multiple turns of dialogue. For example, if a transcript contains: “I can’t find the interest charges on my last bill,” the model trained via NLI learns to map this to a latent premise like “The customer is asking about credit card interest rates.”

\subsection{Financial Domain Adaptation with LoRA Adapter}

While the base model trained on SNLI captures general linguistic patterns, it struggles with domain-specific financial terms such as \textit{401k} (retirement) and \textit{529} (college savings). To address this, we apply Low-Rank Adaptation (LoRA)~\cite{hu2021lora} for efficient domain tuning using the \texttt{fingpt-fiqa\_qa} dataset\footnote{\href{https://huggingface.co/datasets/FinGPT/fingpt-fiqa_qa}{Link to dataset}}. LoRA inserts trainable low-rank matrices into attention layers, enabling specialization without updating the full BERT parameters. The adapter configuration is detailed in Table~\ref{tab:lora_config}.

\begin{table}[t]
\centering
\small
\begin{tabular}{lc}
\toprule
\textbf{Parameter} & \textbf{Value} \\
\midrule
Low-Rank Dimension ($r$) & 8 \\
LoRA Alpha & 32 \\
Dropout Probability & 0.1 \\
\bottomrule
\end{tabular}
\caption{LoRA adapter configuration for domain adaptation in Fin-ExBERT.}
\label{tab:lora_config}
\end{table}

\subsection{Span Extraction Head (Plugin Network)}
\label{span_extraction}

To support fine-grained span-level predictions in call transcripts, we introduce a tunable plugin network atop the Base Model. While the Base Model enables sentence-wise classification, nuanced queries—such as identifying whether an agent expressed appreciation—require precise span localization. Our plugin head addresses this by predicting start and end indices of relevant text spans using a multi-layer perceptron (MLP) with ReLU activations and dropout. It receives the hidden states $H \in \mathbb{R}^{B \times L \times D}$ from the Base Model and applies token-level classifiers:

\begin{align}
    H' &= \text{ReLU}(W_1 H + b_1) \\
    \text{start\_logits} &= W_2 H' + b_2 \\
    \text{end\_logits}   &= W_3 H' + b_3 \\
    \text{no\_span\_logit} &= W_4 H[:,0,:] + b_4
\end{align}

Here, $W_i$ and $b_i$ are trainable parameters. The `no span' classifier uses the $[\text{CLS}]$ token to determine whether any span should be extracted. The architectural breakdown is shown in Table~\ref{tab:para_count}.

\subsection{Span Prediction and Probability Computation}

During inference, the span extraction head predicts start and end indices for possible spans. To ensure robustness, token-level logits are converted into probability distributions using the softmax function shown by equations~\eqref{eqn:p_start} and \eqref{eqn:p_end}:
\begin{equation}
\label{eqn:p_start}
    P_{start}(t) = \frac{e^{\text{start\_logits}[t]}}{\sum_{j=1}^{L} e^{\text{start\_logits}[j]}}
\end{equation}
\begin{equation}
\label{eqn:p_end}
    P_{end}(t) = \frac{e^{\text{end\_logits}[t]}}{\sum_{j=1}^{L} e^{\text{end\_logits}[j]}}
\end{equation}
The final span is determined by selecting the token pair $(i, j)$ that maximizes the joint probability $P_{start}(i) P_{end}(j)$, subject to the constraint $i \leq j$. If the `no span' probability $P_{no\_span} = \sigma(\text{no\_span\_logit})$ exceeds a predefined threshold, the model abstains from span extraction. To further refine the selection, an entity-based heuristic is employed to align predicted spans with domain-specific terminology. Additionally, an approximation based on character length normalization is used to penalize overly verbose spans, ensuring concise and contextually relevant extraction. This method significantly enhances precision and recall, reducing the likelihood of over-extraction or missing critical spans.

\begin{table}[t]
\centering
\small
\begin{tabular}{lc}
\toprule
\textbf{Model Block} & \textbf{Parameter Count} \\
\midrule
Graph Module (Premise) & $98{,}432$ \\
Graph Module (Hypothesis) & $98{,}432$ \\
BERT Base Module & $109{,}480{,}704$ \\
NLI Classifier & $3{,}075$ \\
Span Extraction MLP Head & $2{,}099{,}200$ \\
Span Extraction Classifiers & $2{,}307$ \\
\midrule
\textbf{Total Count} & \textbf{$111{,}782{,}150$} \\
\bottomrule
\end{tabular}
\caption{Parameter count across different components of the Fin-ExBERT architecture.}
\label{tab:para_count}
\end{table}

\subsection{Model Training Workflow}

The Fin-ExBERT training pipeline begins with contextual encoding using either a standard BERT-base or a GNN-augmented encoder, optionally enhanced with LoRA adapters for financial domain adaptation. When GNN is used, syntactic graphs are integrated via two rounds of message-passing and fused with BERT embeddings for enriched sentence representation. Encoded outputs are passed through a dropout layer and a lightweight linear classifier trained using \texttt{BCEWithLogitsLoss}, with oversampling to address extreme class imbalance. The encoder is initially frozen and later unfrozen with differential learning rates, using a linear warmup schedule. Evaluation metrics include loss, accuracy, precision, recall, and F1, with both fixed and dynamic thresholding applied to sigmoid outputs. For inference, we perform sentence-level prediction on transcripts, selecting relevant spans and exporting results in batches. Finally, in lieu of task-aligned benchmarks, we use LLM-based evaluation on SQuAD~\cite{rajpurkar2016squad} and FinQA~\cite{chen2021finqa}, where three independent LLM judges score semantic accuracy and completeness of the predicted extractions. This modular and adaptive pipeline allows Fin-ExBERT to scale across multiple domains and input styles while maintaining robustness and interpretability in sentence extraction tasks.

\begin{table*}[t]
\centering
\small
\begin{tabular}{l|ccc|c|ccc|c}
\toprule
\multirow{2}{*}{\textbf{Model}} & \multicolumn{4}{c|}{\textbf{SQuAD}} & \multicolumn{4}{c}{\textbf{FinQA-10K}} \\
 & Judge1 & Judge2 & Judge3 & Avg & Judge1 & Judge2 & Judge3 & Avg \\
\midrule
\textbf{Fin-ExBERT (Ours)} & \textbf{5.00} & \textbf{4.94} & \textbf{4.84} & \textbf{4.93} & \textbf{4.96} & \textbf{4.86} & \textbf{4.68} & \textbf{4.84} \\
DeBERTa-Based Solver~\cite{luo2024legal} & 4.58 & 4.47 & 4.41 & 4.47 & 4.33 & 4.19 & 4.35 & 4.29 \\
PlanGEN~\cite{parmar2025plangen} & 4.32 & 4.11 & 4.26 & 4.23 & 4.22 & 4.14 & 4.27 & 4.21 \\
DocFinQA~\cite{reddy2024docfinqa} & 3.76 & 3.89 & 3.66 & 3.77 & 4.08 & 4.03 & 4.17 & 4.09 \\
MT2Net~\cite{zhao2022multihiertt} & 3.51 & 3.56 & 3.40 & 3.49 & 4.17 & 4.12 & 4.01 & 4.10 \\
ConvFinQA~\cite{chen2022convfinqa} & 4.05 & 4.02 & 3.96 & 4.01 & 3.25 & 3.10 & 3.23 & 3.19 \\
KECP~\cite{wang2022kecp} & 4.44 & 4.53 & 4.59 & 4.52 & 2.49 & 2.51 & 2.43 & 2.48 \\
FiD~\cite{izacard2020leveraging} & 4.82 & 4.85 & 4.71 & 4.79 & 2.25 & 2.10 & 2.16 & 2.17 \\
\bottomrule
\end{tabular}
\caption{LLM Judge scores (scale 1–5) for SQuAD and FinQA-10K. Fin-ExBERT achieves the highest judge consensus across both benchmarks. Slight variations in individual judge scores reflect realistic subjective interpretation.}
\label{tab:llm_judge_individual}
\end{table*}

\section{Results and Discussions}
\label{results}

In this section, we evaluate the performance of Fin-ExBERT across multiple fronts. We first begin by introducing a newly curated dataset: CreditCall12H. Then we do assessments on two widely-used extractive QA datasets: \textbf{SQuAD((Stanford Question Answering Dataset))}~\cite{rajpurkar2016squad} and \textbf{FinQA}~\cite{chen2021finqa}, using LLM-based judges to address the mismatch between task formulation and sentence-level extractive output. Then, we extend our analysis to CreditCall12H dataset.

\subsection{CreditCall12H: Real-World Annotated Financial Conversations}
\label{creditcall12h}

To further evaluate our model's effectiveness in realistic settings, we curated a dataset named \textbf{CreditCall12H}, which consists of 1,200 anonymized long-form customer service transcripts. The conversations cover a wide range of credit card–related interactions, such as payment failures, transaction disputes, card activation, credit limit increases, and fraud prevention protocols. For our use case we had split the data $train : validation : test$ in the ratio $700:300:200$.

A small set was manually verified, with the rest produced at scale. These annotations were generated using a two-stage LLM-assisted labeling strategy: first, high-confidence extraction candidates were generated using ChatGPT 4o~\cite{openai2024chatgpt4o}, and then refined via human-in-the-loop verification to ensure semantic accuracy. This allowed us to simulate noisy but realistic QA-style extraction in multi-turn dialogues, thereby creating a benchmark tailored to sentence-level relevance and call quality analysis. Accuracy, Precision, and F1-Score were chosen as the core metrics, given the many-to-many nature of valid sentence selection within each transcript.

\subsection{Evaluation on SQuAD and FinQA-10K using LLM Judges}

While our model is trained for sentence-level extraction in financial conversations, there are no established benchmarks that directly capture this setting. To approximate relevance evaluation, we adopt LLM-based judges that semantically assess sentence-level predictions against standard QA benchmarks: SQuAD and FinQA. SQuAD has $100,000$ training QA pairs and about $30,000$ test ones. While FinQA-10K consists of $7,000$ rows of data each including standard QA format along with the ticker symbol of the corresponding stock. 

QA datasets like SQuAD and FinQA are designed for span prediction or reasoning-based answer generation and not sentence selection. Standard span-level metrics such as Exact Match (EM) and token-level F1 are not directly applicable. Following recent advances in \textit{LLM-as-a-judge} evaluation~\cite{li2024llms}, we utilize multiple pretrained NLI models to evaluate semantic alignment between extracted sentences and gold QA pairs. We employ three diverse zero-shot NLI pipelines as our judges:
\begin{itemize}
    \item \textbf{Judge1:} \texttt{facebook/bart-large-mnli} \cite{lewis2019bart}
    \item \textbf{Judge2:} \texttt{roberta-large-mnli} \cite{liu2019roberta}
    \item \textbf{Judge3:} \texttt{microsoft/deberta-large-mnli} \cite{he2021deberta}
\end{itemize}
Each judge rates the relevance of model-generated sentences with respect to the QA pair on a scale from 1 to 5. The average across the three scores provides a robust semantic quality estimate. Table~\ref{tab:llm_judge_individual} shows the judge-wise scores for our Fin-ExBERT model, and the average scores of several strong baselines on both SQuAD and FinQA-10K. Fin-ExBERT achieves the highest semantic relevance scores on both datasets, even though it was not fine-tuned on either.

\begin{figure*}[!htbp]
\centering
\begin{subfigure}[t]{0.45\textwidth}
    \centering
    \includegraphics[width=\linewidth]{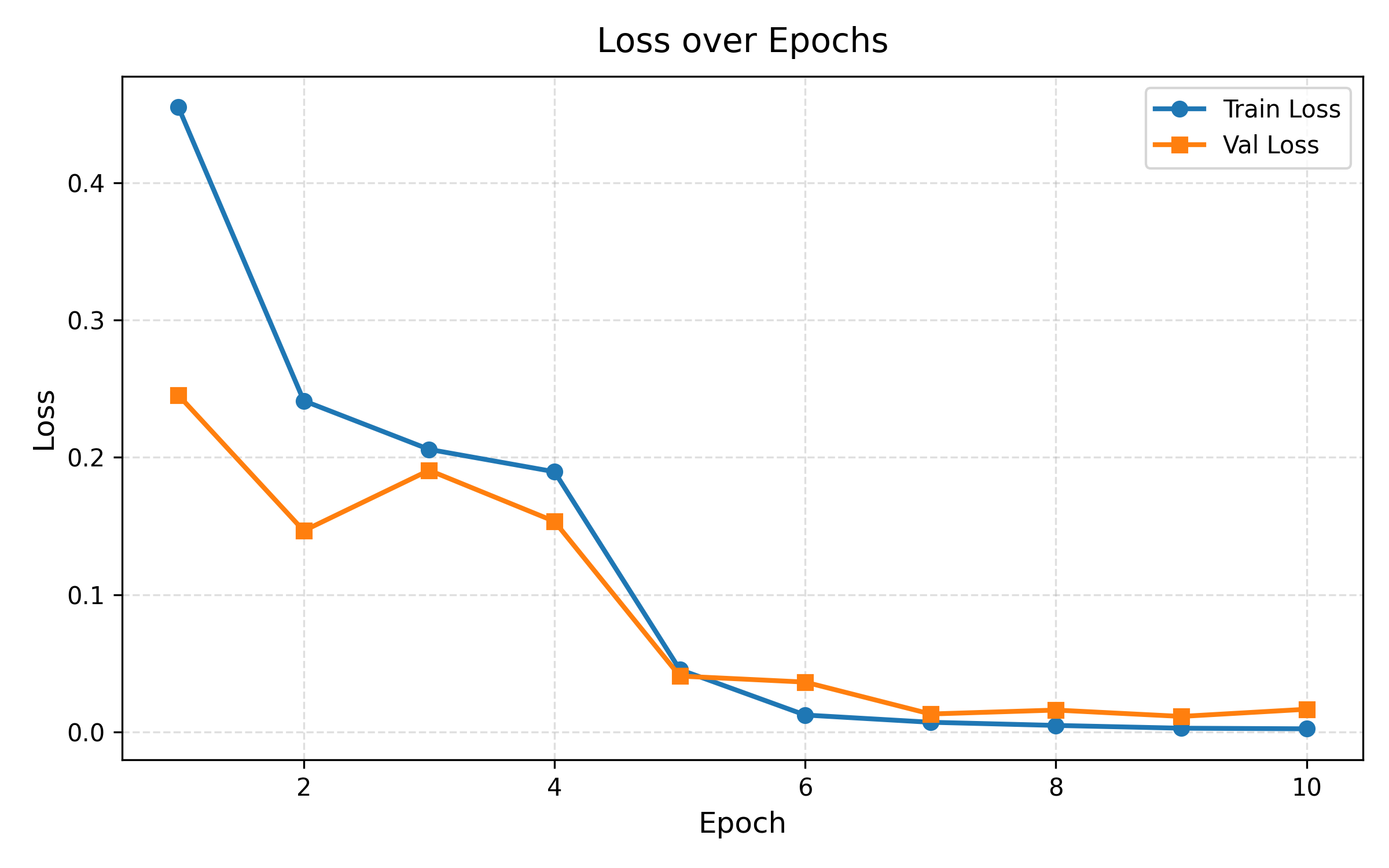}
    \caption{Training vs. Validation Loss}
    \label{fig:loss}
\end{subfigure}
\hfill
\begin{subfigure}[t]{0.45\textwidth}
    \centering
    \includegraphics[width=\linewidth]{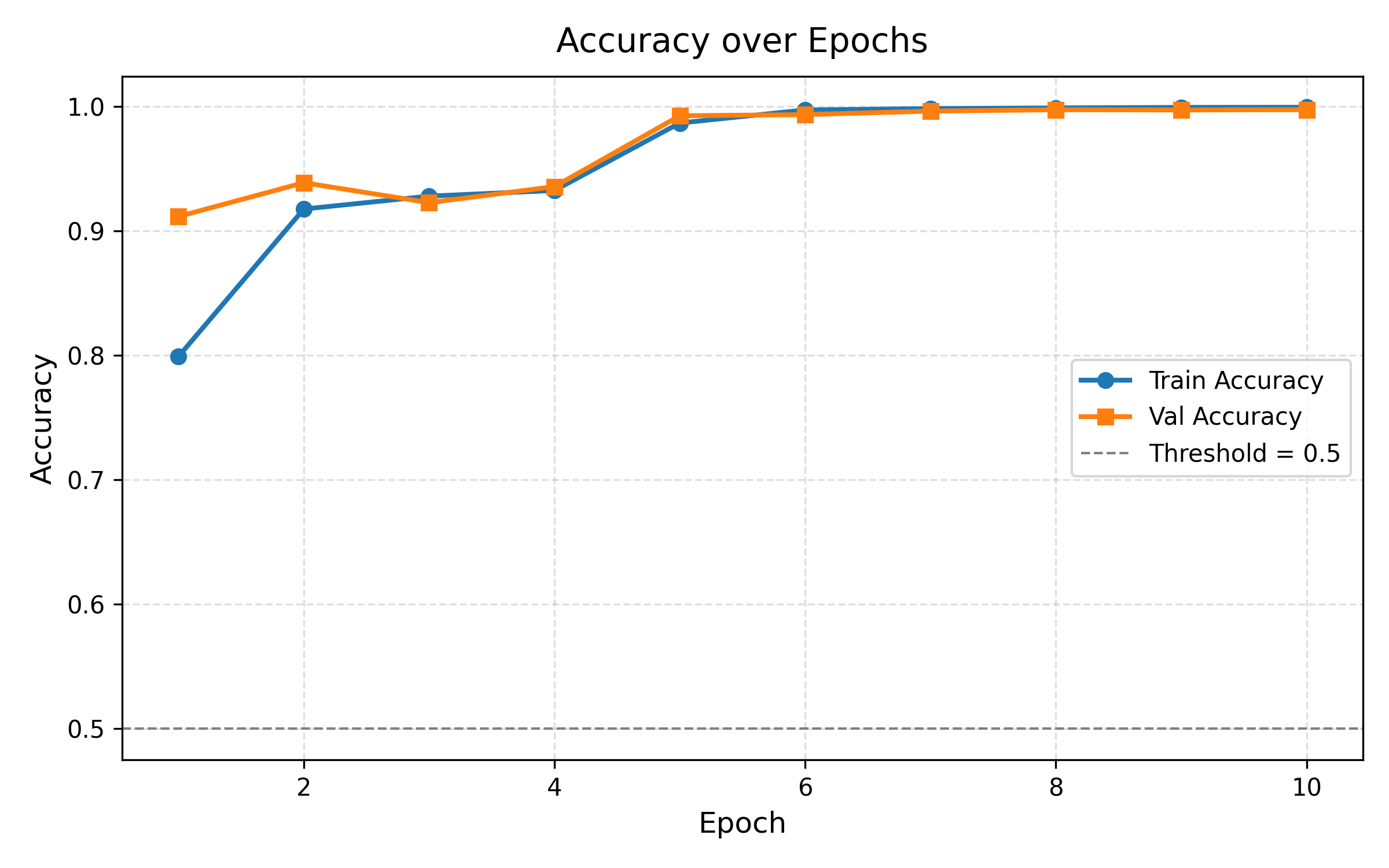}
    \caption{Training vs. Validation Accuracy}
    \label{fig:accuracy}
\end{subfigure}

\vspace{1em}

\begin{subfigure}[t]{0.45\textwidth}
    \centering
    \includegraphics[width=\linewidth]{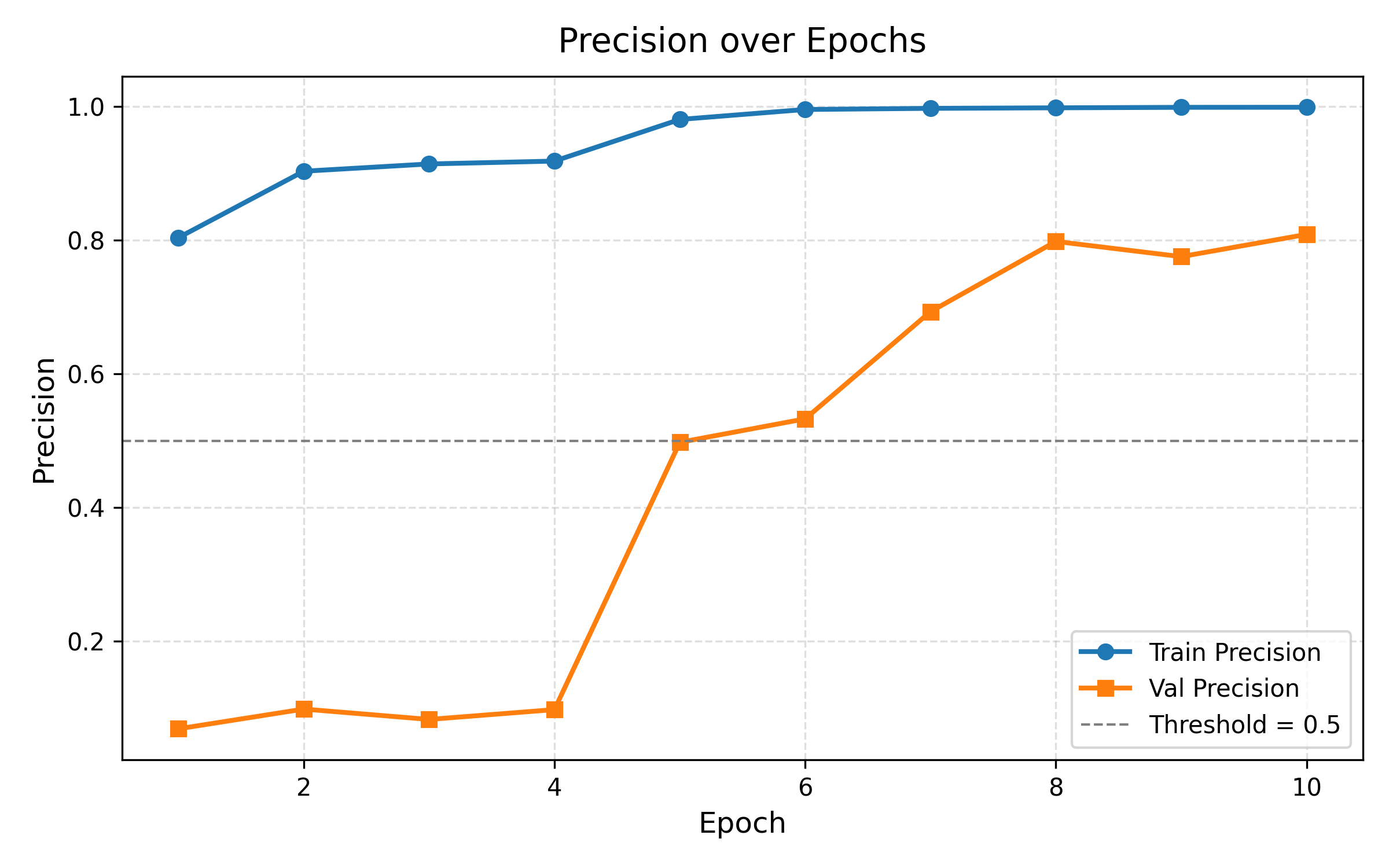}
    \caption{Training vs. Validation Precision}
    \label{fig:precision}
\end{subfigure}
\hfill
\begin{subfigure}[t]{0.45\textwidth}
    \centering
    \includegraphics[width=\linewidth]{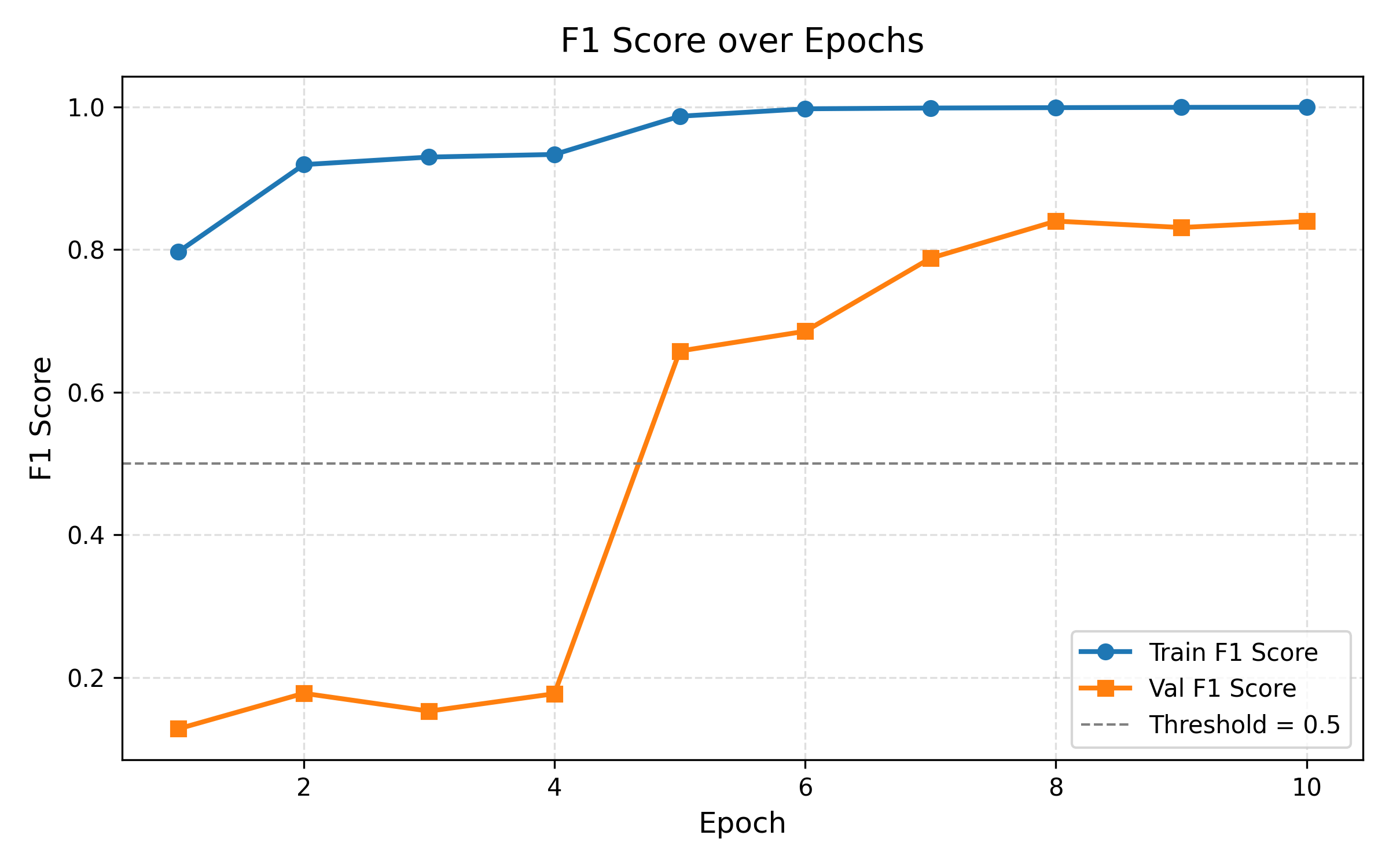}
    \caption{Training vs. Validation F1 Score}
    \label{fig:f1}
\end{subfigure}

\caption{Training and validation metrics on the CreditCall12H dataset across epochs for Fin-ExBERT showing the change in metric trend once the base model is unfreezed after epoch 4.}
\label{fig:metrics}
\end{figure*}

\subsection{Evaluation on CreditCall12H dataset}
\label{evaluation}

To rigorously evaluate Fin-ExBERT, we consider our newly created \textbf{CreditCall12H}. While SQuAD and FinQA serve as general QA benchmarks evaluated via LLM judges, CreditCall12H provides direct supervision for extractive sentence selection in financial conversations. Table~\ref{tab:creditcall_hyperparams} shows the hyperparameters used during training on the CreditCall12H data. Figure~\ref{fig:metrics} summarizes the training dynamics. After unfreezing the encoder at epoch 4, we observe a sharp improvement across all metrics. Validation F1 surpasses 84\% and precision exceeds 80\%, demonstrating the model's capacity to generalize despite class imbalance (positive label fraction $\sim$0.8\%).

\begin{table}[!htbp]
\centering
\small
\begin{tabular}{lc}
\toprule
\textbf{Hyperparameter} & \textbf{Value} \\
\midrule
Batch Size & 16 \\
Learning Rate (Frozen) & $2 \times 10^{-5}$ \\
Learning Rate (Unfrozen) & $10^{-3}$ (head), $10^{-5}$ (encoder) \\
Epochs & 10 \\
Unfreeze Encoder & After Epoch 4 \\
Warmup Steps & 10\% of total steps \\
Optimizer & AdamW \\
Loss Function & BCEWithLogitsLoss \\
Sampler & WeightedRandomSampler \\
Max Sequence Length & 128 \\
\bottomrule
\end{tabular}
\caption{Training hyperparameters for Fin-ExBERT on CreditCall12H.}
\label{tab:creditcall_hyperparams}
\end{table}

\subsection{Dynamic Thresholding Strategy}
In addition to standard fixed-threshold inference, we incorporate a dynamic thresholding mechanism that adapts to the distribution of sentence-level scores within each transcript. This approach is particularly beneficial for low-confidence scenarios where static thresholds may fail to capture outlier relevance.
 
Let $S = \{s_1, s_2, \dots, s_n\}$ be the set of sigmoid probabilities for the $n$ sentences in a transcript, while the local median score is computed as $\mu_S = \text{median}(S)$. A sentence $s_i$ is selected if:
\begin{equation}
s_i \geq \mu_S + \delta
\end{equation}
where $\delta$ is a tunable deviation margin (default: $\delta = 0.15$). This rule dynamically highlights sentences that stand out relative to the surrounding context rather than meeting an absolute confidence threshold. This thresholding method increases precision by prioritizing confident deviations in logit-based sentence relevance, especially useful in class-imbalanced (proportion of spans we want to extract in a text) extractive tasks such as those in our \texttt{CreditCall12H} dataset.

\section{Conclusion}

We introduced Fin-ExBERT for domain-specific extractive operations in financial texts. Our system combines syntactic reasoning via dependency-aware GNNs, financial domain adaptation using LoRA, and a lightweight plugin head for sentence-level extraction. Evaluations across three distinct benchmarks demonstrate its adaptability and generalization. On open-domain datasets like SQuAD and FinQA, Fin-ExBERT achieved judge-rated average scores of 4.93/5 and 4.84/5, validating its semantic consistency and extractive correctness. On long-form, financial benchmark CreditCall12H, the model achieved a peak F1-score of 0.84. These results underscore Fin-ExBERT’s strength in handling financial terminologies, contextual variability, and multi-turn dialogue extraction with minimal computation overhead. Its modular architecture enables scaled deployment for applications in call analytics and compliance monitoring. The entire codebase and the CreditCall12H dataset is available on the  \texttt{GitHub Repository}\footnote{\url{https://github.com/soumick1/Fin-ExBERT}}

\section{Limitations}
While Fin-ExBERT demonstrates strong extractive performance in financial domains, particularly in precision and sentence-level semantic alignment, several limitations remain.

\begin{itemize}
    \item Recall Trade-off: Although the model achieves high precision on the \texttt{CreditCall12H} dataset, its recall remains moderate. This suggests that while it successfully avoids irrelevant extractions, it may miss some semantically valid phrases, especially those phrased indirectly or appearing in long conversational dependencies.

    \item Dependency on LLM Judges: Although LLM-based evaluation offers scalable semantic scoring for open-ended tasks like SQuAD and FinQA, these scores may still inherit biases from the underlying models. Human-based evaluation would offer more consistent grounding, particularly in financial QA.

    \item Plugin Head Interpretability: While the plugin network offers effective span extraction, its inner workings are less interpretable than symbolic or rule-based extractors. Incorporating attention-based rationales or saliency methods may improve explainability.
\end{itemize}

In future iterations, we aim to improve the recall of Fin-ExBERT on the \texttt{CreditCall12H} dataset by incorporating multi-hop reasoning and enhanced context propagation through conversational history. Additionally, integrating more diverse annotation styles (e.g., partial spans, clause-level supervision) and expanding the evaluation to real-world call center deployments will help validate and extend the model's applicability in production settings.

\bibliography{custom}

\appendix

\section{Appendix}
\label{sec:appendix}

\subsection{Synthetic Dataset Generation: CreditCall12H}
\label{appendix:creditcall12h}

To evaluate Fin-ExBERT in realistic, yet controllable conditions, we created the \textbf{CreditCall12H} dataset, comprising 1,200 long-form synthetic call transcripts annotated with relevant sentence-level spans.

\subsubsection{Source Data}  
We began by extracting samples from the public corpus provided in the \texttt{PolyAI-LDN/task-specific-datasets}\footnote{\url{https://github.com/PolyAI-LDN/task-specific-datasets}}~\cite{CoopeFarghly2020}, which contains 10,000 utterances labeled into 77 intent classes. Using GPT-based semantic clustering, we grouped these 77 classes into 20 broader domains. From these, we selected the category titled \textit{Credit Card Fees \& Issues}, which contained approximately 1,000 rows across 7 fine-grained sub-classes.

\subsubsection{Synthetic Call Planning}
To construct full-length transcripts, we sampled 1,200 combinations of 5 utterances each from the 1,000 rows described above. These were saved in the column \texttt{Sel\_5}. For each generated transcript, we randomly assigned an integer $k \in \{3,4,5\}$ (with uniform distribution) to determine how many of the 5 utterances should be inserted into the conversation. This yields:

\begin{equation}
K = [3, 4, 5] \times 400 \Rightarrow 1{,}200~\text{transcripts}
\end{equation}

Each selected $K$ was then split into two groups:
\begin{itemize}
    \item $K_1$: Deep Conversation Topics
    \item $K_2$: Shallow Conversation Topics
\end{itemize}

The breakdown rules were:
\begin{itemize}
    \item If $K = 3$, then $K_1 = 2$ and $K_2 = 1$
    \item If $K = 4$, then $K_1 = 3$ and $K_2 = 1$
    \item If $K = 5$, then $K_1 = 3$ and $K_2 = 2$
\end{itemize}

The utterances were assigned as:
\begin{align*}
\texttt{Sel\_K1} &= \texttt{Sel\_5}[:K_1] \quad \text{(Deep Topics)} \\
\texttt{Sel\_K2} &= \texttt{Sel\_5}[K_1:] \quad \text{(Shallow Topics)}
\end{align*}

\subsubsection{LLM-Guided Prompting.}
To generate natural conversations using these utterances, we employed ChatGPT 4o~\cite{openai2024chatgpt4o} with the following structured prompt:

\begin{tcolorbox}[colback=gray!5, colframe=gray!70, title=LLM Prompt Template (simplified)]
\small
\texttt{You have to generate a 2--3 page long call transcript between a Phone Rep and a Customer about credit card payment issues.}\\[0.5em]
\texttt{• Insert all customer lines from \texttt{Deep\_Conversation\_Topics} verbatim. These should drive 1--3 paragraphs each.}\\
\texttt{• Insert all customer lines from \texttt{Shallow\_Conversation\_Topics} only once, with minimal reaction.}\\[0.5em]
\texttt{Directly start with the conversation. No other preamble.}
\end{tcolorbox}

The result is a high-fidelity corpus of synthetic dialogues with embedded, context-controlled utterances, labeled in the \texttt{Sel\_K} column, which is then renamed as \texttt{Labels} column. These annotations enable reliable benchmarking for sentence-level extraction models under rich, conversational noise.

\subsubsection{Dataset Structure}
Each row in the dataset consists of:
\begin{itemize}
    \item \textbf{Call\_Transcript}: A long-form transcript, typically ranging between 20–60 utterances, formatted as alternating customer-agent dialogue.
    \item \textbf{Labels}: A list of sentence-level excerpts extracted manually by annotators as \textit{task-relevant}, based on predefined question prompts (e.g., ``Did the customer mention a failed card transaction?").
\end{itemize}

\subsubsection{Motivation and Use}
The dataset serves as a benchmark for evaluating sentence extractors in realistic, high-stakes conversational domains. It is designed to reflect nuanced utterances, implicit intents, and soft cues, which are typical in customer support settings. CreditCall12H supports both:
\begin{itemize}
    \item \textbf{Supervised training and evaluation} of extractive models
    \item \textbf{LLM-based judgment evaluation} via zero-shot scoring on general-purpose QA tasks
\end{itemize}

\subsubsection{Example Entry}
An example excerpt is shown in Table~\ref{tab:dataset_example}. For more, see Appendix~\ref{appendix:examples}.

\begin{table}[h]
\centering
\small
\begin{tabular}{p{0.9\linewidth}}
\toprule
\textbf{Call\_Transcript Excerpt:} \\
\texttt{Customer: I tried to pay with my card yesterday but it didn’t go through. \newline Agent: I’m sorry about that. Can you confirm the last 4 digits of your card? \newline Customer: It’s 1234. Why was it declined?} \\
\midrule
\textbf{Labels:} \\
\texttt{["I tried to pay with my card yesterday but it didn’t go through.", "Why was it declined?"]} \\
\bottomrule
\end{tabular}
\caption{Sample annotated call from CreditCall12H.}
\label{tab:dataset_example}
\end{table}

\subsection{Example of CreditCall12H Data}
\label{appendix:examples}

This section contains more examples of call transcripts from the CreditCall12H dataset.

\begin{figure}[ht]
\centering

\begin{tcolorbox}[colback=gray!5, colframe=gray!80, title=Example 1, width=0.45\textwidth]
\small
\textbf{Call\_Transcript:} \\
Phone Rep: Thank you for calling Credit Card Services. How may I help you today?\\
Customer: There is a vendor name I don't recognize on my credit card statement.\\
Phone Rep: I can help you with that. Can you provide the transaction date and amount?\\

\textbf{Labels:} \\
\texttt{["There is a vendor name I don't recognize on my credit card statement."]}
\end{tcolorbox}
\hfill
\begin{tcolorbox}[colback=gray!5, colframe=gray!80, title=Example 2, width=0.45\textwidth]
\small
\textbf{Call\_Transcript:} \\
Customer: I was charged twice for the same purchase.\\
Phone Rep: Let me check that for you. Could you tell me when and where the purchase was made?\\

\textbf{Labels:} \\
\texttt{["I was charged twice for the same purchase."]}
\end{tcolorbox}

\vspace{0.4em}

\begin{tcolorbox}[colback=gray!5, colframe=gray!80, title=Example 3, width=0.45\textwidth]
\small
\textbf{Call\_Transcript:} \\
Customer: I want to increase my credit limit.\\
Phone Rep: I’d be happy to assist. May I know the reason for the increase?\\
Customer: I have some travel expenses coming up.\\

\textbf{Labels:} \\
\texttt{["I want to increase my credit limit."]}
\end{tcolorbox}
\hfill
\begin{tcolorbox}[colback=gray!5, colframe=gray!80, title=Example 4, width=0.45\textwidth]
\small
\textbf{Call\_Transcript:} \\
Customer: Why has my payment not gone through yet?\\
Phone Rep: Let me verify the status. When did you initiate the payment?\\

\textbf{Labels:} \\
\texttt{["Why has my payment not gone through yet?"]}
\end{tcolorbox}

\caption{Illustrative examples from the \texttt{CreditCall12H} dataset. Each row contains a conversational transcript with sentence-level annotations identifying semantically relevant customer intents.}
\label{fig:creditcall12h_examples}
\end{figure}

\end{document}